\documentclass{article}

\usepackage{arxiv}

\usepackage[utf8]{inputenc} 
\usepackage[T1]{fontenc}    
\usepackage{hyperref}       
\usepackage{url}            
\usepackage{booktabs}       
\usepackage{amsfonts}       
\usepackage{nicefrac}       
\usepackage{microtype}      
\usepackage{lipsum}
\usepackage{graphicx}
\usepackage{subfig}
\usepackage{amsmath}

\title{Using neuroevolution for designing soft medical devices}

\author{
    Hugo Alcaraz-Herrera\\
    Unconventional Computing Laboratory, \\
    College of Arts, Technology and Environment, \\
    University of the West of England, \\
    Bristol, BS16 1QY, United Kingdom \\ \texttt{hugo.alcaraz@uwe.ac.uk}
\And
    Michail-Antisthenis Tsompanas  \\
    Unconventional Computing Laboratory \& \\
    School of Computing \& Creative Technologies,\\
    College of Arts, Technology and Environment, \\
    University of the West of England,\\ 
    Bristol, BS16 1QY, United Kingdom \\
    \texttt{antisthenis.tsompanas@uwe.ac.uk}
\And
       Andrew Adamatzky \\
       Unconventional Computing Laboratory,\\
    College of Arts, Technology and Environment, \\
    University of the West of England,\\ 
    Bristol, BS16 1QY, United Kingdom \\
\And
       Igor Balaz\\
       Laboratory for Meteorology, Physics and Biophysics,\\ Faculty of Agriculture, \\
       University of Novi Sad, \\ Trg Dositeja Obradovica 8, 21000, Novi Sad, Serbia
}

\begin{document}
\maketitle

\begin{abstract}
Soft robots can exhibit better performance in specific tasks compared to conventional robots, particularly in healthcare related tasks. However, the field of soft robotics is still young, and designing them often involves mimicking natural organisms or relying heavily on human experts' creativity. A formal automated design process is required. The use of neuroevolution-based algorithms to automatically design initial  sketches of soft actuators that can enable the movement of future medical devices, such as drug-delivering catheters, is proposed. The actuator morphologies discovered by algorithms like Age-Fitness Pareto Optimisation, NeuroEvolution of Augmenting Topologies (NEAT), and Hypercube-based NEAT (HyperNEAT) were compared based on the maximum displacement reached and their robustness against various control methods. Analyzing the results granted the insight that neuroevolution-based algorithms produce better-performing and more robust actuators under diverse control methods. Specifically, the best-performing morphologies were discovered by the NEAT algorithm.
\end{abstract}


\section{Introduction}\label{sec:introduction}

Although soft robotics is a promising research area with extensive applicability, it entails considerable challenges. One of the most intricate challenges is discovering a proper design: finding the most suitable robot design requires substantial time and material resources, because, it implies numerous prototypes being tested in real life \cite{Schulz2016}. Unfortunately, for design of soft robots this is not the only complex aspect, since they use flexible materials whose mechanical properties are non-linear and complex to characterize \cite{Hiller2014}. For instance, {\em biohybrid machines}, a particular type of soft robots, are built employing biological components, such as cells or tissues, which are inherently difficult to control with conventional methods. This brings another set of challenges, since several parameters need to be considered during the design process and the reality gap problem is becoming more profound.

One promising strategy to counteract these intrinsic challenges of soft robots is using NeuroEvolution (NE). This methodology was previously utilised to design reconfigurable biohybrid machines without strict definitions of specific targets \cite{Kriegman2020}. NE consists of evolving the topology and weights of artificial neural networks (ANNs) through a genetic algorithm (GA). One of the most effective and well-known approaches is NeuroEvolution of Augmenting Topologies (NEAT) \cite{Stanley2002}, which has demonstrated an adequate performance regarding morphology generation \cite{Auerbach2011}. Moreover, NEAT was extended under the rationale that natural structures are composed of patterns and shape repetition. This extension is called Hypercube-based NeuroEvolution of Augmenting Topologies (HyperNEAT) \cite{Stanley2009}. This approach evolves a particular type of networks, i.e., Compositional Pattern-Producing Networks (CPPNs), which use, among others, periodic functions, such as sine and cosine, to generate symmetry and patterns that help to evolve topologies \cite{Stanley2007}.

The primary objective of this research is to design {from scratch} soft actuators that can be implemented in medical devices, such as catheters for targeted drug delivery in areas of the human body that are otherwise difficult to reach. In order to define a trustworthy design engine, NEAT and HyperNEAT are validated as tools to design soft actuator morphologies. We chose the catheter scenario as an exemplary case study for automated morphology generation due to its straightforward design. {The inputs for the proposed methodology are the physical characteristics of two different types of material used as building blocks and the constraints that the actuator morphology needs to follow to resemble a catheter actuator. The fitness function of the possible solutions considered is the maximum bending angle that the actuator morphology can reach \textit{in silico}. The output of the evolutionary methodologies is the best combination of the two predefined material building blocks that are composed together inside the predefined design space to produce the actuator morphology.} This streamlined focus on bending, which allows other automated systems to concentrate on optimising flexibility, ensuring that generated models are both functional and adaptable to various medical scenarios.

{NE algorithms were utilised here to evolve the design of soft actuators morphologies under an indirect representation of the three-dimensional space. Specifically, instead of using a bitstream of all possible dimensional segments to encode the characteristics of every point in the available design space (i.e. the type of material used for that point), the NE evolves the artificial networks that when queried with all the points in the design space, they return the material characteristic for each point of the morphology design.}

The validation methodology implemented in this research compares the performance of NEAT and HyperNEAT against Age-Fitness Pareto Optimisation (AFPO), an evolutionary algorithm designed to avoid premature convergence \cite{Hornby2006}. Three metrics are used to assess the performance of the algorithms: (i) the general performance in terms of the displacement observed in the free end of the actuator during a specific time-frame; (ii) the robustness of the designs of actuators produced; and (iii) the total volume of actuators. The fitness of the possible solutions is provided by a physics engine simulating the actuator. The aforementioned metrics and an updated set of constraints applied in the physics engine, present an altered search space for the algorithms which proved to be efficient in previous works of designing soft and biohybrid robots tested on purely the longer distance covered in unconstrained locomotion tasks \cite{Alcaraz2024,Kriegman2020}.

{Whereas the methodology of NE in optimisation of morphologies of soft robots and biohybrid machines was previously studied \cite{Kriegman2020,Tanaka2022,Matthews2023}, the contribution of this work is towards targeting a specific real-world application. Namely, previous works used evolution towards unconstrained locomotion with a generic fitness function (i.e., the longest point reached by a morphology and some variations of that but without any constraints), the present study describe an environment with specific constraints, in order to better reflect the application under study. In specific, here one end of the morphology is fixed, to simulate the insertion point of a drug-delivery catheter. Moreover, while previous works \cite{Kriegman2020} used AFPO with mutation only operators, here we use NEAT and HyperNEAT with mutation and recombination operators to explore in more efficient way the design space of the problem.}

The rest of the paper is organized as follows. Section~\ref{sec:background} describes background research focusing on soft robot related tasks and in specific utilising NEAT and HyperNEAT algorithms. Section~\ref{sec:algorithms} introduces in detail {the automated design methodology and the} mechanisms of the three algorithms utilised for experimentation in this research.  Section~\ref{sec:catheter} describes the experiments performed in the task of designing soft actuators for catheters. Finally, Section~\ref{sec:conclusions} concludes this work by presenting insights that emerged from the research and outlines future work. 


\section{Background}\label{sec:background}

The concept of {\em soft robots} can be traced back to Isaac Asimov's books published in the 1940s-1950s \cite{Asimov1950}. However, the first scientific and engineering papers related to soft grippers~\cite{hirose1978development} and soft architecture machines~\cite{negroponte1976soft} started in the late 1970s. The domain has started to flourish in late 1980s with further publications on soft fingers~\cite{akella1989manipulating,bicchi1990intrinsic,chen1982gripping}, electrostatic actuators~\cite{gillespie1985optical}, and fluid based grippers~\cite{kenaley1989electrorheological}. Unlike traditional and rigid robots, these robots are built utilising flexible and ductile materials \cite{Lipson2014, Rus2015}. Soft robots have the potential to exhibit better performance in specific tasks that are related to search and rescue~\cite{milana2022soft,murphy2000marsupial,wang2019novel,wen2024design}, space exploration~\cite{methenitis2015novelty,aracri2021soft,ng2023untethered,mintchev2018soft}, and healthcare~\cite{patino2016miniaturized,majidi2014soft,hsiao2019soft} due to their mechanisms and morphology, which are inspired by the movement and behavior of living organisms. 

Publications on soft robotic catheters, and relevant surgical devices, started to emerge from late 1980s~\cite{kwoh1988robot} and by 2020 there was a range of catheters developed for cardio-vascular surgery~\cite{nguyen2023development,rogatinsky2023multifunctional,nguyen2022bidirectional}, urinary system~\cite{levering2014soft,baburova2023magnetic,yang2022magnetic}, endovascular treatment~\cite{gopesh2021soft}, \textit{in situ} bioprinting~\cite{zhou2021ferromagnetic}, endoscopy~\cite{li2022soft} and drug delivery~\cite{hu2024catheter}. Plantoid robots proposed by Mazzolai and colleagues~\cite{mazzolai2014emerging,mazzolai2018robots,friedman2021interview,mazzolai2017can,mazzolai2010plant,sadeghi2016plant} can be considered relevant to the research on robotic catheters. These robots are soft-bodied mechanical devices capable of physically growing at the tip (thanks to embedded 3D printing techniques), similar to a plant root, and can change the direction of their growth due to incorporated environmental condition sensors.

{The use of soft robots in medical education is also noteworthy. For instance, a recent work \cite{peng2024peristaltic} implements McKibben artificial muscle on a soft robot inspired by the large intestine and demonstrates similar behavior to its biological counterpart. This soft robot was designed and 3D printed in a prototype that was tested along with a simulation of finite element analysis to validate its performance. As a result, this soft robot may be employed by medical students to improve their understanding of the functionality of large intestine via optical interaction and touch. Moreover, the same robotic architecture was considered with an updated control mode to achieve higher speeds and loads of object interaction \cite{peng2024controlling}. Drawing inspiration from the movement of inchworm, rather than the intestine peristalsis, enhanced its transport capabilities. The investigation of several of the design and control aspects provided significant insights into the enhanced efficiency and adaptability of its application on object transport tasks.}

Several research studies have been conducted to find suitable strategies for soft robot design. For instance, in \cite{Bhatia2021}, a large-scale benchmark platform for optimising the design and control of soft robots is introduced. The software is called {\em Evolution Gym}, and robots can be composed of different types of voxels such as soft, rigid actuators. The benchmark environments contained in Evolution Gym provide a wide range of tasks, such as locomotion and object manipulation. Robots are designed in a two-dimensional layout.

Furthermore, another design tool for multi-material soft actuators, {\em SoRoForge}, is introduced in \cite{Smith2023}. This platform comprises the major stages of this type of actuator's design, evaluation, and fabrication process. The software allows: (i) an interactive exploration of implicit geometry functions represented by computational networks; (ii) seeing an online three-dimensional static geometry preview of designs; and (iii) the creation of three-dimensional printable design files.

Another design automation system for soft actuators is presented in \cite{Smith2022}. The design engine is the well-known multi-objective evolutionary approach called {\em Non-dominated Sorting Genetic Algorithm-II} (NSGA-II) \cite{Deb2002} that was implemented to find suitable morphologies of soft actuators. Results suggest that under this approach, it is necessary to define adequate fitness functions to find suitable morphologies of soft robots since the ones found in the series of experiments described in this work did not outperform those solutions that were {\em a priori} manually designed and were used as a baseline.

Evolutionary approaches are not the only methods employed to design soft robots. For instance, in \cite{Matthews2023}, a gradient-based approach, called {\em novo optimisation}, is utilised to find suitable structures of robots. The algorithm is tested using terrestrial locomotion tasks and shown to be capable of: (i) assessing the fitness of the behavior of the soft robot; (ii) identifying deficiencies in its overall shape, topology, number and shape of limbs, mass distribution, musculature, and behavioral control; and (iii) simultaneously changing all these aspects which leads to an improvement in evaluation.

In the same path of non-evolutionary approaches, a study introduces a soft robot design methodology based on topology optimisation \cite{Zhang2017}. The authors implement their methodology to design a soft gripper capable of undergoing free travel and delivering a grasping force. Moreover, the design can be printed using three-dimensional additive manufacturing. However, the optimisation model assumes the homogeneity of materials.

Regarding the two neuroevolution-based algorithms studied here, NEAT has been implemented in numerous studies focusing on soft robots. An example of these studies presents a three-link planar arm model driven by nine muscles, whose purpose is to simulate the human arm \cite{Wen2017}. A neurocontroller trained by NEAT controls the muscular-skeleton arm. The training sets applied to NEAT are built considering the forward kinematics, geometry relationships, and muscle mechanic equations. NEAT is compared against a fixed-topology ANN. Results indicate that the neurocontroller trained by NEAT outperforms the traditional ANN in critical aspects such as moving the arm to a specific position and converging the distance between the endpoint and the target point with a minimum distance error.

NEAT has not only been studied for controlling bioinspired mechanisms. It also has been utilised as a design engine for soft robots. For instance, in \cite{Cheney2015}, NEAT, alongside CPPNs, is utilised to design soft robots capable of reaching or squeezing through a small aperture. The experiments consisted of simulations where each soft robot is placed within a cage whose dimensions are $15 \times 15 \times 11$ voxels (arbitrary volume in the physics simulator), leaving a gap of 1 voxel in the $x$ and $y$ axes between the edge of the cage and the maximum size of the soft robot: $11 \times 11 \times 11$. The cage is rigid, immobile and indestructible. Each side has an opening, approximately a circle of diameter 10. Although this research is primarily considered as a proof-of-concept, it is demonstrated that more compliant and deformable soft robots are more capable of reaching or squeezing through small apertures, such as tunnels, than soft robots that are less flexible. 

Following the same path of being utilised as morphology generator, in \cite{Auerbach2010}, NEAT (and CPPNs) is used to produce three-dimensional physical structures capable of conserving momentum to achieve maximum displacement due to gravity. Structures are composed of spherical cells which fuse to make rigid bodies. The growth process commences with a single cell called {\em root} and is located at the origin of the structure. A cloud of $n$ points is set around the cell with the $n$ points distributed on the root's surface. Once the cloud is over the cell, every point is used to query a CPPN whose output represents the concentration of matter at that point. A matter threshold determines where to allocate a cell: the more the output value exceeds the matter threshold, the denser the cell allocated at that point will be. Results suggest that the proposed method can produce artefacts that suitably capture the non-obvious relationship between function and physical structure.

Another study where NEAT helps to design morphologies of soft robots capable of navigating across environments with different levels of viscosity is described in \cite{Jha2018}. The approach consists of evolving ANNs that observe the virtual environment and respond to it by controlling the muscle force of the soft robot. The performance of morphologies is tested in three different environments: (i) low-viscosity drag; (ii) intermediate-viscosity drag; and (iii) high-viscosity drag. Results indicate that the properties of the environment have a strong influence on soft robot design. For instance, it is possible to observe that the evolutionary pressure flats the parts with big surfaces of the morphologies, which exploits the viscosity drag~\cite{Jha2018}.

NEAT has not only been used for soft robots focused on one task. In \cite{Kimura2021}, this method is utilised to build morphologies of soft robots capable of performing more than one specific task. The study argues that by combining the genotype of soft robots focused on single-functional tasks is feasible to generate multi-functional robots. The experiments involved in this study consisted of simulating voxel-based creatures performing tasks in terrestrial and aquatic environments. Results suggest that the proposed method: (i) enables the efficient exploration of the morphology search space; (ii) can find morphologies that satisfy two tasks simultaneously faster than existing methods. 

On the other hand, HyperNEAT has been primarily utilised in tasks targeting the control of robots. An example of these studies presents a method where morphologies are the input, and the output is neural network controllers capable of working on these different morphologies \cite{Risi2013}. Each neural controller is tested in three different quadruped morphologies whose lengths are 0.25, 0.30, and 0.37 meters. The neural controllers generated by HyperNEAT are compared against static controllers. Results indicate that HyperNEAT can identify the relationship between morphology and controller architectures during evolution, which leads to a suitable performance.

In the same direction of controller design, in \cite{Clune2009} legged robots are studied, and particularly four-legged robots. The rectangular torso of the robot morphology is 0.15 wide, 0.3 long, and 0.5 tall (in arbitrary ODE physics simulator units). Each leg has three cylinders (radius 0.02, length 0.075) and three hinge joints. HyperNEAT is compared against {\em Fixed-Topology} NEAT (FT-NEAT), also known as Perceptron NEAT. Results indicate that HyperNEAT outperforms FT-NEAT since its morphologies: (i) vastly outperform those generated by FT-NEAT in every generation and (ii) exhibit better coordination than those created by FT-NEAT. Another relevant insight of this research is that generative encodings can outperform direct encodings on regular problems.

Another study, described in \cite{Yosinski2011}, focuses on creating gaits for quadruped robots. HyperNEAT is compared against six parameterized learning strategies: (a) uniform random hill climbing; (b) Gaussian random hill climbing; (c) policy gradient reinforcement learning; (d) Nelder-Mead simplex; (e) a random baseline; and (f) a new method that builds a model of the fitness landscape with linear regression. Nine servos actuate the robot used for experimentation, each commanded to a position in the $[0, 1023]$ range, corresponding to a physical range of $[-120^\circ,+120^\circ]$. Two on-board batteries power the servos. Results advocate the superiority of HyperNEAT over all parameterized learning strategies since the gaits it generates exhibit significantly better performance.

Not only standalone neurocontrollers can be evolved through HyperNEAT. In \cite{DAmbrosio2008}, it is stated that it is possible to evolve neurocontrollers encoded by a single genome representing a team of predator agents that work together to capture prey. In the experiments conducted, agents are not aware of their teammates, and due to prey avoidance from nearby predators, it is feasible for one predator to undermine another's pursuit by knocking its prey off its path. Consequently, predators are forced to learn coherent roles that complement the roles of their allies. Results indicate that encoding a team of agents as a unified genome brings benefits such as: (i) critical skills do not need to be rediscovered for separate agents; (ii) due to CPPNs representing multiagent policies, they are assigned to separate agents as a function of their relative geometry and simultaneously exploiting the internal geometries of agents. 

HyperNEAT has been scarcely used to design morphologies of soft robots. For instance, in \cite{Tanaka2022}, HyperNEAT is employed to evolve controllers and morphologies simultaneously. Robots are evaluated considering their capabilities to adapt in four different scenarios: (i) moving in a flat terrain in a limited time; (ii) moving in a highly uneven terrain; (iii) climbing through a narrow stepwise channel; and (iv) throwing a solid box. Results suggest that the method generates robots that show a suitable performance in all the aforementioned tasks.

More recently, another study tests the suitability of HyperNEAT (and NEAT) as a design engine for soft robot morphologies, which is described in \cite{Alcaraz2024}. The experiments focused on locomotion tasks, particularly reaching the maximum displacement possible in 10 seconds. Morphologies generated by AFPO are considered as a baseline. Results indicate that both neuroevolution-based approaches can outperform AFPO, and HyperNEAT can perform more suitably than NEAT, despite the absence of geometrical aspects of the problem domain in the design of the substrate. This confirms what is argued in \cite{Clune2009Sensitivity}: HyperNEAT can obtain suitable results, even if there is no defined geometrical representation of the problem. 

Thus, based on the aforementioned research findings, it is feasible to consider NEAT and HyperNEAT as the core strategy to design adequate morphologies for soft actuators for catheters. {Furthermore, considering the three-dimensional nature of the morphologies, Voxelyze is a suitable software to simulate their physical behavior. Although Voxelyze has limited capacity in capturing complicated physical properties (i.e., does not take into account fluid dynamics), it was selected as an efficient alternative against more complicated simulators. The basic functionality (i.e., interactions between adjacent voxels, volumetric actuation and collision detection) is considered adequate under the trade off between a lower reality gap and a computational efficient evolutionary optimisation. Nonetheless, all candidate solutions in the evolutionary optimisation will undergo the same inaccuracies imposed by the simulator. 
Moreover, Voxelyze has demonstrated adequate accuracy in simulating biohybrid machines \cite{Kriegman2020} that exhibit inherently high levels of behavioral uncertainty. This has been achieved with only one round of fine-tuning based on \textit{in vitro} observations, aligning simulations more closely with physical prototypes.}

{In the medical domain, however, the accuracy required for final designs is paramount, as high fidelity in biohybrid actuator performance is essential. Therefore, before advancing any selected design to manufacturing prototyping, these preliminary designs should undergo a secondary stage of rigorous testing in simulations that account for more precise physics-based properties, including fluid dynamics and nuanced biological interactions. By coupling our approach with a more detailed simulator at next developmental stage, we can better approximate real-world performance, mitigating risks associated with physical prototyping and ensuring that any final designs meet the strict reliability and performance standards needed for medical applications.}

{Some previous studies focus in a two-dimensional layout design of soft robots \cite{Bhatia2021,Tanaka2022}, the present work aims towards more realistic three-dimensional designs, thus, Voxelyze is utilised. Whereas some non-evolutionary approaches were previously proposed for soft robot design \cite{Matthews2023,Zhang2017}, the utilisation of evolutionary methods is preferred here, due to their simplicity and enhanced capacity to avoid local optima in rugged optimisation problems. Nonetheless, the direct representation of individuals in evolutionary soft robot design techniques has proved not as efficient as generative representation (i.e. CPPNs) \cite{cheney2014unshackling}. The implementation of AFPO has proved very efficient when employed in the designing process of biohybrid machines \cite{Kriegman2020}. However, it was previously applied using only a mutation operator, without crossover, which limited the method’s exploratory capacity. The authors of the prior study recommended incorporating recombination-based methods, such as NEAT and HyperNEAT, for future improvements. Finally, the implementation of NE algorithms have been tested in specific applications, namely achieving the longest distance on an unconstrained plane or with some obstacles located on the plane and object manipulation \cite{cheney2014unshackling,Kriegman2020,Tanaka2022,Matthews2023}. Here, we study the application of NE algorithms on a more reserved and specialized application (namely, the angular movement of a soft actuator enabling a medical catheter).}


\section{Methodology}\label{sec:algorithms}

{The design of soft actuators in a given three-dimensional space ($[x,y,z]$) was produced automatically via an evolutionary process over an indirect representation and evaluated through a physics engine (i.e. Voxelyze) to fit the initially set target. The target (or the fitness function) here was set to achieve the higher possible bending angle, since, the application target is a drug-delivering catheter. To represent the insertion point of the catheter, one end of the morphology was fixed and not able to move, thus, emulated an anchoring point for the actuator. Moreover, instead of using a direct representation in the evolutionary optimisation for the design of the soft actuator, namely a one-to-one correspondence of all possible positions in the three-dimensional design space with their physical characteristics, an indirect representation was preferred. The indirect representation used in all of the algorithms described in the following, is constituted by neural networks, that have as input all the possible combination of the three-dimensional space ($[x,y,z]$) and as outputs the physical characteristics of the material comprising the actuator morphology in the corresponding area.} 

{The fitness function consists of a physics engine called {\em Voxelyze} (also available in \cite{reconf}), which simulates the physical response of morphologies under determined conditions. Morphologies are defined within Voxelyze by stacking voxels (the minimum building block in Voxelyze) of different materials together. Here, two types of voxels were considered for the simplicity of the designing landscape: one active and one passive voxel, which can represent the response of contractile muscles and immotile cells, respectively. The physics simulation parameters for both these building blocks were set to Young’s modulus of $5*10^6$ Pa, Poisson's Ratio of $0.35$, and coefficients of static and dynamic friction of $1.0$ and $0.5$, respectively. Moreover, the active voxel is characterized by volumetric actuation of $\pm50\%$ of the volume at rest at $4Hz$. These parameters were selected based on previously published and \textit{in vitro} validated works \cite{Kriegman2020}, but do not limit the generalization of the algorithmic methodologies.}

{The original version of Voxelyze was modified (changes in the source code available in \cite{tsomCode}) to trace the position of the catheter actuator in the $x$,$y$,$z$ axes during a time $t$. Furthermore, one end of actuators is fixed in such a way that they only move vertically (i.e., one degree of freedom) in the $yz$ plane. In addition, actuators have a passive enclosure to follow the findings of previous simulation works \cite{tsompanas2024outline} and constraints of laboratory experiments, i.e. the bioreactor that a muscle actuator requires for the supply of nutrients in case of a biohybrid machine. The output of Voxelyze contains three values that describe the position of the actuator free tip (with respect to $x$,$y$,$z$ dimensions) for the initial position and the final one (after the predefined simulation time), and the number of voxels that compose the actuator.}

Three algorithms were utilised in this research. Namely, Age-Fitness Pareto Optimisation (AFPO), NeuroEvolution of Augmenting Topologies (NEAT), and Hypercube-based NeuroEvolution of Augmented Topologies (HyperNEAT), which will be described in Section~\ref{sec:algorithms_afpo}, Section~\ref{sec:algorithms_neat}, and Section~\ref{sec:algorithms_hyperneat} respectively.

\subsection{AFPO}\label{sec:algorithms_afpo}

This approach was conceived as an alternative method to avoid premature convergence in some evolutionary algorithms \cite{Schmidt2010}. One of the fundamental aspects of AFPO is {\em age}, which, under this scope, can be understood as how long genotypic material has existed in the population \cite{Hornby2006}. Based on this, the age of a solution (i.e., an individual) is measured in generations.

The algorithm commences initializing the individuals composing the population at random, with age of all individuals set to one. If an individual survives to the next generation, its age is incremented by one. When genetic operators (i.e., crossover and mutation) take place, the age is inherited as the maximum age of the parents. 

\begin{figure}[tb!]
  \centering
     \includegraphics[width=0.57\linewidth]{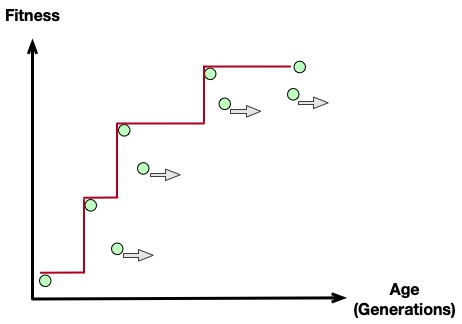}
  \caption{An example of a population moving in a two-dimensional Pareto space where the vertical axis is fitness and the horizontal axis is age. Figure adapted from \cite{Schmidt2010}.}
  \label{fig:algorithms_afpo}
\end{figure}

Since AFPO employs a single population, it tracks each individual's fitness similarly to a standard evolutionary algorithm, including the genotypic age. This multi-objective optimisation procedure aims to determine the non-dominated Pareto front of the problem domain, where the objective is maximizing fitness with minimum age. One way of visualizing the mechanism of AFPO is the population (i.e., individuals) evolving in a two-dimensional plane of age and fitness. Figure~\ref{fig:algorithms_afpo} illustrates this dynamic. 

\subsection{NEAT}\label{sec:algorithms_neat}

This algorithm was designed to counteract on three main issues that previous NE algorithms exhibit: (i) a solution representation capable of allowing arbitrary topologies to recombine; (ii) avoiding the premature disappearance of topological novelties discovered throughout the evolutionary process; and (iii) avoiding complex topologies without using a fitness function focused on punishing topological complexity \cite{Stanley2002}.

The first issue was tackled using {\em gene tracking} by {\em historical markings}. This is achieved with a list of connection genes which represent a connection between two nodes containing: (a) the ``origin'' node and the ``destination'' node, (b) the weight of the connection, and (c) a Boolean state of the connection (enabled or disabled). Moreover, each gene has one {\em global innovation number}, a unique numerical ID and the core of the crossover operator. 

\begin{figure}[tb!]
  \centering
     \includegraphics[width=1.0\linewidth]{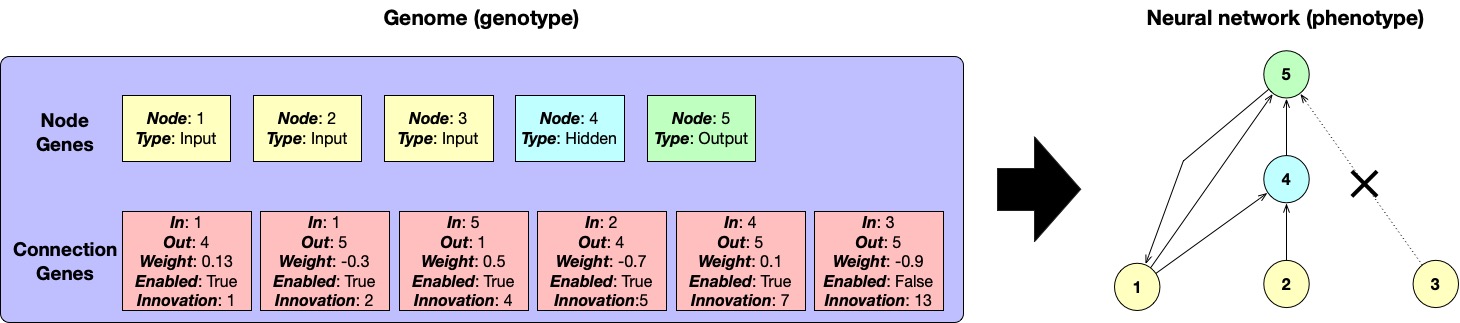}
  \caption{Example of a genotype and its phenotype under NEAT. Figure adapted from \cite{Stanley2002efficient}.}
  \label{fig:algorithms_neat_genotype}
\end{figure}

Two genes having the same global innovation number (also known as ``historical origin'') represent the same structure; therefore, the genes of both genomes that have the same innovation number are aligned. The genes that do not match are inherited from the fittest genome or chosen at random. When a new gene is generated by mutation, the global innovation number increases and is associated with the new gene. Innovation numbers are generally the chronology of every gene during the evolutionary process. Figure~\ref{fig:algorithms_neat_genotype} presents an example of a genotype (i.e., a neural network) and its associated phenotype. 

The second issue, was addressed by {\em speciation to protect innovation}, which generates species within the population aiming at the interaction of individuals that belong to the same species. NEAT creates species utilising topological similarities that are identified by the use of historical markers. The canonical way to measure compatibility of two genomes (i.e., individuals) is by the number of excess and disjoint genes. Thus, the less disjoint genes are, the more compatible the two individuals are. The compatibility distance ($\delta$) between two genomes is as follows \cite{Stanley2002efficient}:

\begin{equation}\label{eq:neat_distance} 
    \delta = \frac{c_{1}E}{N} + \frac{c_{2}D}{N} + c_3 \times \overline{W}
\end{equation}

\noindent
where $E$ is the number of excess genes, $D$ represents the disjoint genes, $\overline{W}$ is the mean weight differences of matching genes, $c_1$, $c_2$, $c_3$ are utilised to determine the relevance of $E$, $D$, and $\overline{W}$, respectively. Regarding $N$, it represents the number of genes in the individual with the largest number of genes; it is used to normalize the genome size during the evolutionary process. The distance $\delta$ determines in which species an individual is allocated utilising a compatibility threshold $\delta_t$. Individuals are allocated into the first species (taken at random) where their distance $\delta$ is $<\delta_t$, assuring that individuals do not belong to more than one species.

NEAT evaluates individuals utilising {\em explicit fitness sharing}. Under this schema, individuals within the same species share the fitness, which implies a limitation for species to expand over the population \cite{Goldberg1987}. Since this mechanism adjusts the fitness of individuals by dividing an original fitness by individuals in each species, the species grow or shrink if their average adjusted fitness is either over or under the population average. 

The third issue was alleviated by {\em minimizing network structures}, which is feasible due to the mechanism of speciation to protect innovation. Since NEAT can initialize individuals whose input neurons are fully connected to the output neurons (i.e., no hidden nodes are involved), new structures are incrementally generated when structures are mutated during the evolutionary process. The structures that exhibit a suitable performance can survive. The effect induced by this approach narrows down the search towards a minimal number of weight dimensions, which implies: (a) a reduction of runtime to find solutions \cite{tsompanas2024incremental} and (b) avoiding the generation and evaluation of complex structures during the evolutionary process.

\subsection{HyperNEAT}\label{sec:algorithms_hyperneat}

\begin{figure}[tb!]
  \centering
     \includegraphics[width=0.83\linewidth]{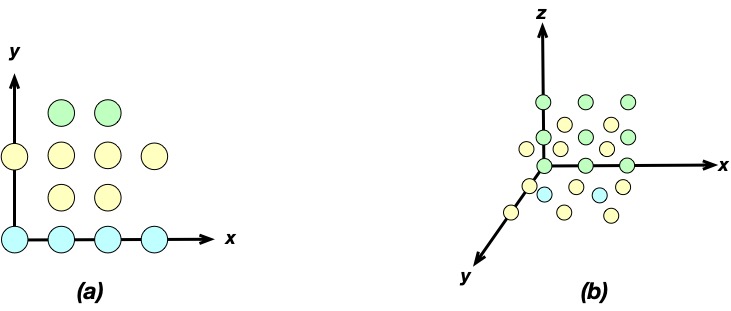}
  \caption{Two typical configurations of substrates utilised in HyperNEAT: (a) two-dimensional, and (b) three-dimensional.}
  \label{fig:algorithms_hyperneat_substrate}
\end{figure}

This approach is an extension of NEAT \cite{Stanley2009}, which utilises the evolutionary engine of the aforementioned algorithm to evolve a particular class of neural networks known as CPPNs \cite{Stanley2007}. CPPNs are used due to their ability to generate patterns, such as repetition and symmetry. There exist two key differences between NEAT and HyperNEAT:

\begin{enumerate}

    \item {\em Activation functions}. Typically, NEAT generates ANNs containing hidden nodes exclusively using the sigmoid function. HyperNEAT, nevertheless, allows the use of other activation functions, such as Gaussian, trigonometric, and periodic, in the nodes composing CPPNs. In this way, CPPNs can evolve, exploring a significantly more expansive search space of network functionality.
    
    \item {\em Substrate}. HyperNEAT is capable of embodying the geometry of the domain problem due to the nature of CPPNs; therefore, the topology of ANNs is computed, taking into account their geometry. Thus, the geometric layout where HyperNEAT operates is called a {\em substrate}, which can be configured in diverse manners. One of the most popular configurations consists of a set of nodes allocated in a two-dimensional plane, known as {\em grid}. Another widely used configuration is the {\em three-dimensional grid}, which contains a set of nodes allocated in a three-dimensional space. Figure~\ref{fig:algorithms_hyperneat_substrate}-a depicts an example of a grid, whereas Fig.~\ref{fig:algorithms_hyperneat_substrate}-b presents an example of a three-dimensional grid.       
\end{enumerate}

Under HyperNEAT, topologies of ANNs embodied in the substrate are generated by CPPNs employing the position of the neurons. Consequently, when a two-dimensional substrate is used, the weight between the neuron $m$ and the neuron $h$ is calculated using their position in the substrate:

\begin{equation}\label{eq:hyperneat_2d_weight} 
    CPPN(x_m,y_m,x_h,y_h) = weight_{mh}
\end{equation}

\noindent
Regarding the bias of neurons, they are computed, providing the coordinates of the neuron in the ``origin'' position, whereas the ``destination'' coordinates are set to zero. Thus, the bias of neuron $m$ is obtained by:

\begin{equation}\label{eq:hyperneat_2d_bias} 
    CPPN(x_m,y_m,0,0) = bias_{m}
\end{equation}

\noindent
Furthermore, if a three-dimensional grid is being utilised, the weight between the neuron $h$ and the neuron $m$ is computed as follows:

\begin{equation}\label{eq:hyperneat_3d_weight} 
    CPPN(x_h,y_h,z_h,x_m,y_m,z_m) = weight_{hm}
\end{equation}

\noindent
The bias of neurons is calculated analogously to the manner described in Equation~\ref{eq:hyperneat_2d_bias}. Therefore, the bias of neuron $h$ is calculated by:

\begin{equation}\label{eq:hyperneat_3d_bias} 
    CPPN(x_h,y_h,z_h,0,0,0) = bias_{h}
\end{equation}

In general, the mechanism of HyperNEAT starts by defining the substrate. The next step consists of generating the initial population of CPPNs. Then, for each CPPN, the connections (i.e., weights) and bias of the substrate are calculated using Eqs.~\ref{eq:hyperneat_2d_weight} and \ref{eq:hyperneat_2d_bias} for a two-dimensional grid, and Eqs.~\ref{eq:hyperneat_3d_weight} and \ref{eq:hyperneat_3d_bias} for a three-dimensional grid. Once the substrate is built, its performance (and hence the performance of the CPPN) is evaluated, and a fitness score is assigned to the CPPN. After the evaluation stage, NEAT is applied to evolve the population of CPPNs. It is important to emphasize that the outputs of CPPNs are usually in the $[-1.0,1.0]$ range, which, in case it is necessary, implies a normalization or mapping procedure.
 
Summarizing, HyperNEAT aims to evolve the topology (i.e., the connectivity) and weights of ANNs that are represented by a substrate with a specific geometry through CPPNs.


\section{Experimental setup and Results}\label{sec:catheter}

In this research, three metrics are used to determine the suitability to generate soft actuator morphologies of NEAT and HyperNEAT: (i) the general performance in terms of the displacement of the actuator in the $yz$ plane during a simulated time period; (ii) the robustness of the fittest actuators found; and (iii) the number of voxels composing the actuator. Furthermore, AFPO (see Section~\ref{sec:algorithms_afpo}) is utilised as a baseline actuator generator in these experiments, and the implementation employed is available on GitHub \cite{reconf} and is fully described in \cite{Kriegman2020}.

\subsection{Experimental setup}\label{sec:catheter_setup}

The population is composed of 100 individuals (i.e., CPPNs) for the three approaches. Each evolutionary run lasted 3000 generations. The set of activation functions  utilised for experimentation is the same as used in  \cite{Kriegman2020}, and is composed of: {\em sigmoid}, {\em sine}, {\em negative sine}, {\em square}, {\em negative square}, {\em square root of absolute}, {\em negative square root of absolute}, {\em absolute}, and {\em negative absolute}. Activation functions are randomly chosen during CPPN initialization and mutations. In order to perform a fair comparison for all approaches, the implemented initialization procedure of the population is described in \cite{Kriegman2020}. Moreover, for NEAT and HyperNEAT, the parameters employed for the evolutionary process of CPPNs are shown in Table~\ref{tab:catheter_parameters}. For each approach, 20 independent experimental runs were performed, by changing only the initial random populations.

\begin{table}
\caption{Parameters utilised to evolve CPPNs under NEAT and HyperNEAT for the actuator generator domain.}
\label{tab:catheter_parameters}
 \begin{center}
  \begin{tabular}{ c c } 
   \toprule
   Parameter & Value \\
   \midrule
   compatibility threshold & 3 \\
   compatibility disjoint coefficient & 1.0 \\ 
   compatibility weight coefficient & 0.5 \\
   maximum stagnation & 25 \\
   survival threshold & 0.6 \\ 
   activation function mutate rate & 0.4 \\
   adding/deleting connection rate & 0.3/0.2 \\
   activating/deactivating connection rate & 0.5\\
   adding/deleting node rate & 0.3/0.2 \\
   \bottomrule
  \end{tabular}
 \end{center}
\end{table}

Regarding the dimensions of actuators, in terms of voxels, there are 20 units in the $x$ axis and 8 units in the $y$ and $z$ axes. Thus, the axes range of the three-dimensional layout where morphologies are designed are as follows: for $x$ axis, $[0,20]$. For $y$ and $z$ axes, $[0,8]$.

{Since simulations require a significant amount of computational time, it is essential to mitigate this limitation. Thus, the implementation related to AFPO was previously written under the multiprocessing paradigm \cite{Kriegman2020} and was used here under the same principle. The implementation for NEAT and HyperNEAT was designed under a client-server architecture to take advantage of distributed computing capacities \cite{Alcaraz2024}. The hardware utilised for experimentation in this study is described as follows: {\em Processor}: ARM (virtualised), nine cores (18 threads), 3.20GHz, {\em RAM Memory}: 16GB, LPDDR5. Provided that the evolutionary optimisation is implemented under a client-server architecture, the utilisation of High-performance computing (HPC) can significantly accelerate this methodology, so that it complies with time constraints of real-world optimisation problems. Additionally, some alternations in the algorithmic implementation, like the initialization methodology, have proved to contribute with an acceleration of 30\% on the computations \cite{tsompanas2024incremental}.}

\subsubsection{NEAT configuration}\label{sec:catheter_setup_neat}

Due to actuators being designed in a discrete three-dimensional layout, it is necessary to provide for each point across the layout: (i) the presence or not of a material voxel, and (ii) the material type of each voxel. Thus, under NEAT, CPPNs are queried as follows:

\begin{equation}\label{eq:catheter_neat} 
    CPPN(x_i,y_i,z_i) = v_i,m_i
\end{equation}

\noindent
where the tuple $(x_i,y_i,z_i)$ represents the coordinates of the $i$-$th$ point in the three-dimensional layout, $v_i$, refers to the presence of a voxel in the $i$-$th$ point of the layout, and it is processed as the presence or absence of a voxel by the following equation:

\begin{equation}\label{eq:catheter_neat_v}
    Presence(x_i,y_i,z_i) = \begin{cases}
    \text{yes,  } & |v_i| \geq 0.5 \\ 
    \text{no, } & \text{otherwise}
    \end{cases}
\end{equation}

\noindent
Furthermore, $m_i$ represents the type of material of the voxel in the $i$-$th$ point of the layout. In the scope of this research, two types of materials are considered: (i) {\em passive}, encoded as 1, and (ii) {\em contractile}, encoded as 3. Equation~\ref{eq:catheter_neat_m} presents the mechanism to map $m_i$ as an encoded material ID.

\begin{equation}\label{eq:catheter_neat_m}
    Material~code(x_i,y_i,z_i) = \begin{cases} 
    1\text{,  } & |m_i| < 0.5 \\ 
    3\text{,  } & \text{otherwise}              
    \end{cases}
\end{equation}

\subsubsection{HyperNEAT configuration}\label{sec:catheter_setup_hyperneat}

\begin{figure}[tb!]
  \centering
     \includegraphics[width=0.57\linewidth]{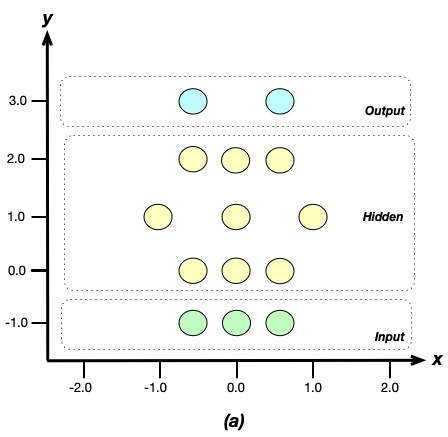}
  \caption{Substrate employed under HyperNEAT for actuator generator.}
  \label{fig:catheter_hyperneat_substrate}
\end{figure}

The first element of HyperNEAT to determine is the substrate. Thus, based on the discrete three-dimensional layout used to design actuators, the substrate has three neuron inputs. Furthermore, since the data required to determine are the presence of a voxel in the $i$-$th$ point and the material of that voxel, the amount of output neurons is two. Equation~\ref{eq:catheter_hyperneat} presents how substrates are queried.

\begin{equation}\label{eq:catheter_hyperneat} 
    substrate(x_i,y_i,z_i) = v_i,m_i
\end{equation}

\noindent
where the tuple $(x_i,y_i,z_i)$ is the coordinates of the $i$-$th$ point in the three-dimensional layout. The $v_i$ variable is associated with the presence or absence of a voxel in the $i$-$th$ point of the layout and is mapped using Eq.~\ref{eq:catheter_neat_v}, and $m_i$ is related to the type of the voxel material in the point $i$ of the layout and is processed by Eq.~\ref{eq:catheter_neat_m}. Figure~\ref{fig:catheter_hyperneat_substrate} depicts the design of the two-dimensional substrate used for experimentation. The neuron allocation used for the substrate was decided by conducting a series of experiments varying the number of neurons per hidden layer and the number of hidden layers in the $[1,5]$ range. Furthermore, the activation function implemented in the substrate is {\em SELU} due to its fast convergence and self-normalizing properties \cite{Klambauer2017}.

Once the substrate is defined, the following step defines how CPPNs are queried. Since the substrate was defined in a two-dimensional space, CPPNs are queried utilising Eq.~\ref{eq:hyperneat_2d_weight} for calculating the weights among neurons, and Eq.~\ref{eq:hyperneat_2d_bias} to compute the bias of the neurons allocated in the substrate. It is essential to point out that based on the definition of HyperNEAT \cite{HyperNEAT2015}, determining whether a connection is calculated is by a threshold. Thus, if $|weight_{mh}| > 0.2$, it is normalized in the $[-3.0,3.0]$ range. Otherwise, there is no connection between the neuron $m$ and the neuron $h$.

\subsection{Analysis of the performance}\label{sec:catheter_analysis}

In order to validate the suitability of catheter actuators generated by AFPO, NEAT and HyperNEAT, three experiments are conducted: (i) analysing the general performance in terms of finding the actuator with the maximum displacement possible (i.e., the fittest) in the $yz$ plane; (ii) testing the robustness of the fittest actuators found; and (iii) identifying the fittest actuators with the minimum volume (i.e., the minimum number of voxels) possible. It is essential to highlight that during all experimentation stages, the displacement was measured in terms of the length of voxels.

\subsubsection{Finding the fittest actuator}\label{sec:catheter_analysis_general}

One vital feature of actuators is the displacement in the $yz$ plane, since they have one end fixed. This displacement can be understood as an indicator of the bending capacity exhibited by actuators: the more displacement the actuator reaches, the more bendable the catheter is. 

For instance, if a soft actuator exhibits no displacement despite a determined stimulation, the angular range where it can operate is $[0, \alpha]$, where $\alpha=0^\circ$, which can be interpreted as the soft actuator presents a poor performance. On the other hand, if the movement observed in the soft actuator allows to satisfy $\alpha>0^\circ$, then the soft actuator can operate in a more comprehensive angular range (i.e., a better performance). In general, the bigger the displacement of the soft actuator (and hence $\alpha$) is, the more comprehensive the angular range where the soft actuator can operate.

In this experiment, the evaluation of actuators is as follows: for each independent experimental run, 25 different controller strategies (or phase offsets in the oscillations of the simulated active voxels) are generated at random and do not change throughout the evolution. For each morphology of the population, the 25 controller phase offsets previously generated are used to stimulate the displacement of the actuator. In other words, each morphology is stimulated 25 times with different controller phase offsets. This procedure occurs for each generation until the experimental run is complete. Thus, the aptitude $apt$ of the actuator $a$ is calculated as follows:

\begin{equation}\label{eq:catheter_analysis_aptitude} 
    apt_a= \frac{\sum_{i=1}^{25} displacement_i}{25}
\end{equation}

\noindent
It is important to emphasize that this experiment does not consider the number of voxels as a penalty for the aptitude value.

\begin{figure}[tb!]
  \centering
     \includegraphics[width=0.67\linewidth]{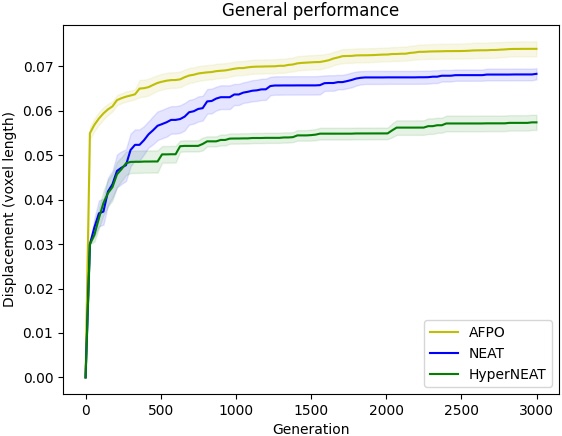}
  \caption{Mean performance of AFPO, NEAT, and HyperNEAT showing $\pm$95\% confidence interval (shaded region).}
  \label{fig:catheter_analysis_general_performance}
\end{figure}

Figure~\ref{fig:catheter_analysis_general_performance} shows the mean performance in finding the fittest individual (i.e., the fittest actuator) with 95$\%$ confidence intervals depicted by the shaded regions across 20 experimental runs under AFPO, NEAT and HyperNEAT. In general, AFPO exhibits a significantly better performance than both neuroevolution-based approaches. Furthermore, NEAT performs better than HyperNEAT. All the data gathered were tested and are not normally distributed (Shapiro-Wilk test; $p<0.01$). Using the Kruskal-Wallis test, it is feasible to confirm that significant differences exist among the performance of the three approaches ($p<0.01$). Therefore, a ranking in terms of performance can be performed: AFPO $>$ NEAT $>$ HyperNEAT (Dunn's test: $p<0.01$).

These results suggest that under 25 different controller phase offsets, AFPO can find the fittest actuators morphologies than those found by NEAT and HyperNEAT. Moreover, HyperNEAT's unexpected performance may be influenced by the absence of domain geometric elements that can be embodied in the substrate design.

\subsubsection{Exploring the robustness of the fittest actuator }\label{sec:catheter_analysis_robutsness}

A catheter could be suitable to deliver drugs in numerous angles under a specific scenario (i.e., a specific controller phase offset of each active voxel composing the actuator). However, it might not exhibit similarly high performance if the scenario changes. This experiment focuses on testing the robustness (i.e., how suitable the performance is regardless of the external conditions) of the three fittest actuators found across the 20 experimental runs conducted using AFPO, NEAT and HyperNEAT.

Each top actuator generated by all approaches is tested utilising 1000 different controller phase offsets of each active voxel. In other words, the controller phase offset scenarios were randomly generated and used by all the morphologies studied in this experiment (i.e., the three fittest actuator morphologies generated by each approach). Thus, the robustness (i.e., the aptitude $apt$) of the actuator $a$ can be measured as follows:

\begin{equation}\label{eq:catheter_analysis_robustness} 
    apt_a= \frac{\sum_{i=1}^{1000} displacement_i}{1000}
\end{equation}

\begin{figure}[tb!]
  \centering
     \includegraphics[width=0.99\linewidth]{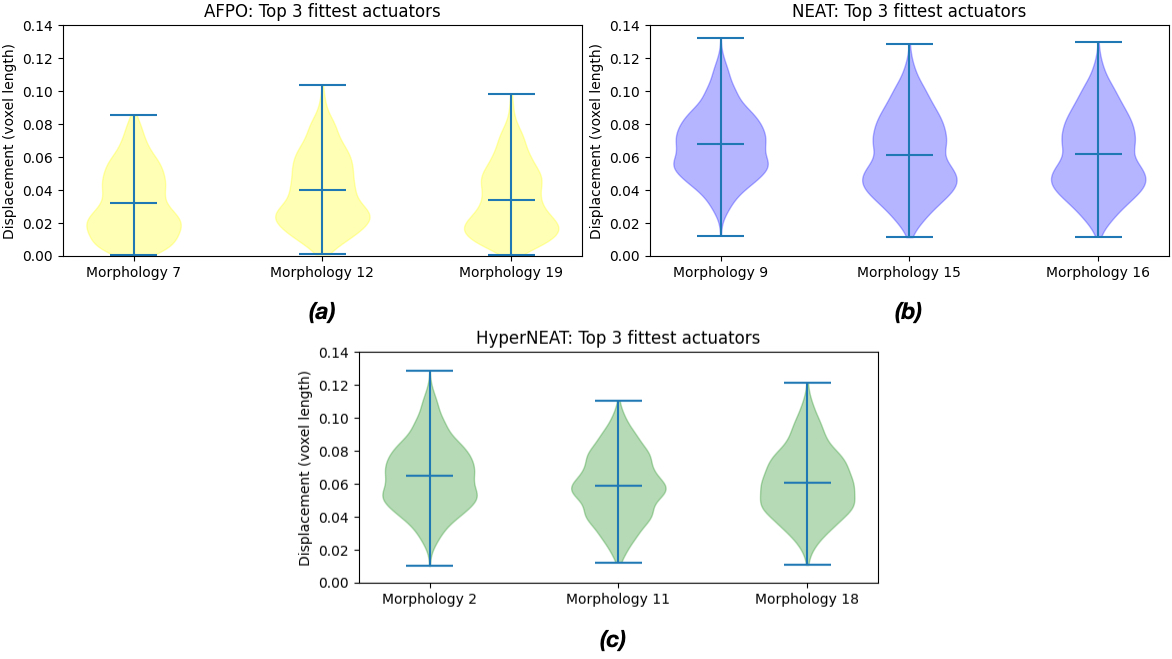}
  \caption{Displacement observed in the $yz$ plane under 1000 different controller phase offset controllers of the top three fittest actuators found in 20 different experimental runs under: (a) AFPO; (b) NEAT; and (c) HyperNEAT.}
  \label{fig:catheter_analysis_robustness}
\end{figure}

Figure~\ref{fig:catheter_analysis_robustness} presents violin plots comparing the performance (i.e., the displacement observed in the $yz$ plane of the free end of the morphology) of the three fittest actuators found by AFPO, NEAT and HyperNEAT. Each violin plot exhibits the maximum, minimum, median and kernel density estimation of the frequency distribution of values under 1000 different offset controller scenarios.

When AFPO morphologies are considered (see Fig.~\ref{fig:catheter_analysis_robustness}-a), all morphologies exhibit significant differences, a fact that is determined by testing all the data collected, and realizing that they are not normally distributed (Shapiro-Wilk test; $p<0.01$). Then by the Kruskal-Wallis test, it is possible to confirm significant differences among the displacement observed in the three actuators ($p<0.01$). Thus, a performance ranking can be performed: Morphology 12 $>$ Morphology 19 $>$ Morphology 7 (Dunn's test: $p<0.01$).

Under NEAT, all morphologies exhibit significant differences, and the data gathered are non-normal (Shapiro-Wilk, Kruskal-Wallis; $p<0.01$). Based on the previous result, it is feasible to produce a ranking of the performance of the three fittest actuators generated by NEAT: Morphology 9 $>$ Morphology 16 $>$ Morphology 15 (Dunn's test: $p<0.01$).

On the other hand, when HyperNEAT results are analysed, the three fittest morphologies generated show significant differences and non-normality is observed in the data gathered (Shapiro-Wilk, Kruskal-Wallis; $p<0.01$). Therefore, a ranking focused on performance can be assumed: Morphology 2 $>$ Morphology 18 $>$ Morphology 11 (Dunn's test: $p<0.01$).

\begin{figure}[tb!]
  \centering
     \includegraphics[width=0.67\linewidth]{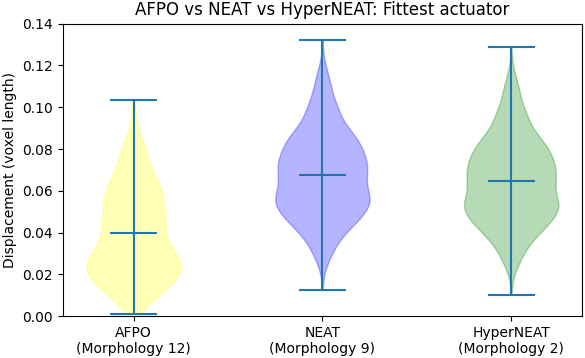}
  \caption{Fittest actuator under: AFPO (left), NEAT (centre), and HyperNEAT (right).}
  \label{fig:catheter_analysis_robustness_full}
\end{figure}

To conclude, Fig.~\ref{fig:catheter_analysis_robustness_full} shows violin plots comparing the performance of the fittest actuator morphology produced by AFPO, NEAT and HyperNEAT. Each violin presents the kernel density estimation of the frequency distribution, maximum, minimum and median values of the 1000 controller phase offset scenarios. All the morphologies present significant differences, confirmed through the Kruskal-Wallis test ($p<0.01$). Based on this result, it is possible to rank the performance of the actuators as follows: NEAT $>$ HyperNEAT $>$ AFPO (Dunn's test: $p<0.01$).

In general, results suggest that neuroevolution-based approaches, such as NEAT and HyperNEAT, are more adept at producing bendable actuators than those generated by AFPO, regardless of the controller phase offset scenario. This realization comes as a contradiction to the previous test, where AFPO seemed as the most appropriate method, whereas, only 25 phase offsets were tested during the optimisation evolutionary process. This implies that NEAT and HyperNEAT have the potential to create more robust actuators than AFPO. Furthermore, despite the comparable performance of NEAT and HyperNEAT, the former has the edge in producing more robust and, consequently, more bendable actuators. {As mentioned previously, HyperNEAT can prove advantageous over problems that have a strong geometric regularity, based on its ability to evolve network structures over geometric substrates. On the other hand, the NEAT algorithm provided better results in this robustness analysis. This can be attributed to two factors. Firstly, the HyperNEAT substrate was only fine-tuned in a limited range of (i) hidden layers and (ii) neurons per hidden layer (i.e., [1,5] for both), because of the high computational requirement related with testing higher ranges. As a result, the best substrate configuration for the given problem can be more complicated than the ones tested within this study. Secondly, NEAT is based on evolving CPPNs (CPPN-NEAT), that have been proved to be very efficient in emerging behaviors that can facilitate movement \cite{cheney2014unshackling}. }

\subsubsection{Identifying the minimum volume of the fittest actuators}\label{sec:catheter_analysis_voxels}

Since the morphologies analysed in this research should inform the design of devices in real life whose objective is to manoeuvre within narrow parts of the human body (e.g., blood vessels), the scale of the manufacturing must be significantly minuscule. From this point of view, the fewer voxels the actuator has, the more suitable the catheter employing that actuator. Table~\ref{tab:voxels_afpo_neat_hyperneat} presents the number of voxels (passive and contractile) composing the fittest actuators produced by AFPO, NEAT, and HyperNEAT.

Even though AFPO produced the actuator with fewer voxels, its performance is less suitable than the approaches based on NE. In contrast, HyperNEAT produced an actuator whose performance is acceptable; however, the number of voxels, particularly the contractile, is significantly high. Regarding the actuator generated by NEAT, it presents a suitable trade-off between performance and the number of voxels since its performance is the best among the three, and the number of voxels is lower than the number of voxels of the actuator designed by HyperNEAT.

\begin{table}
\caption{Number of passive and contractile voxels of the fittest morphologies produced by AFPO, NEAT, and HyperNEAT.}
\label{tab:voxels_afpo_neat_hyperneat}
 \begin{center}
  \begin{tabular}{c c c c} 
   \toprule
   Approach & Total voxels & Passive voxels & Contractile voxels \\
   \midrule
   AFPO & 900 & 632 & 268  \\
   NEAT & 1063 & 632 & 431 \\
   HyperNEAT & 1253 & 632 & 621 \\
   \bottomrule
  \end{tabular}
 \end{center}
\end{table}

\begin{figure}[tb!]
  \centering
     \includegraphics[width=0.93\linewidth]{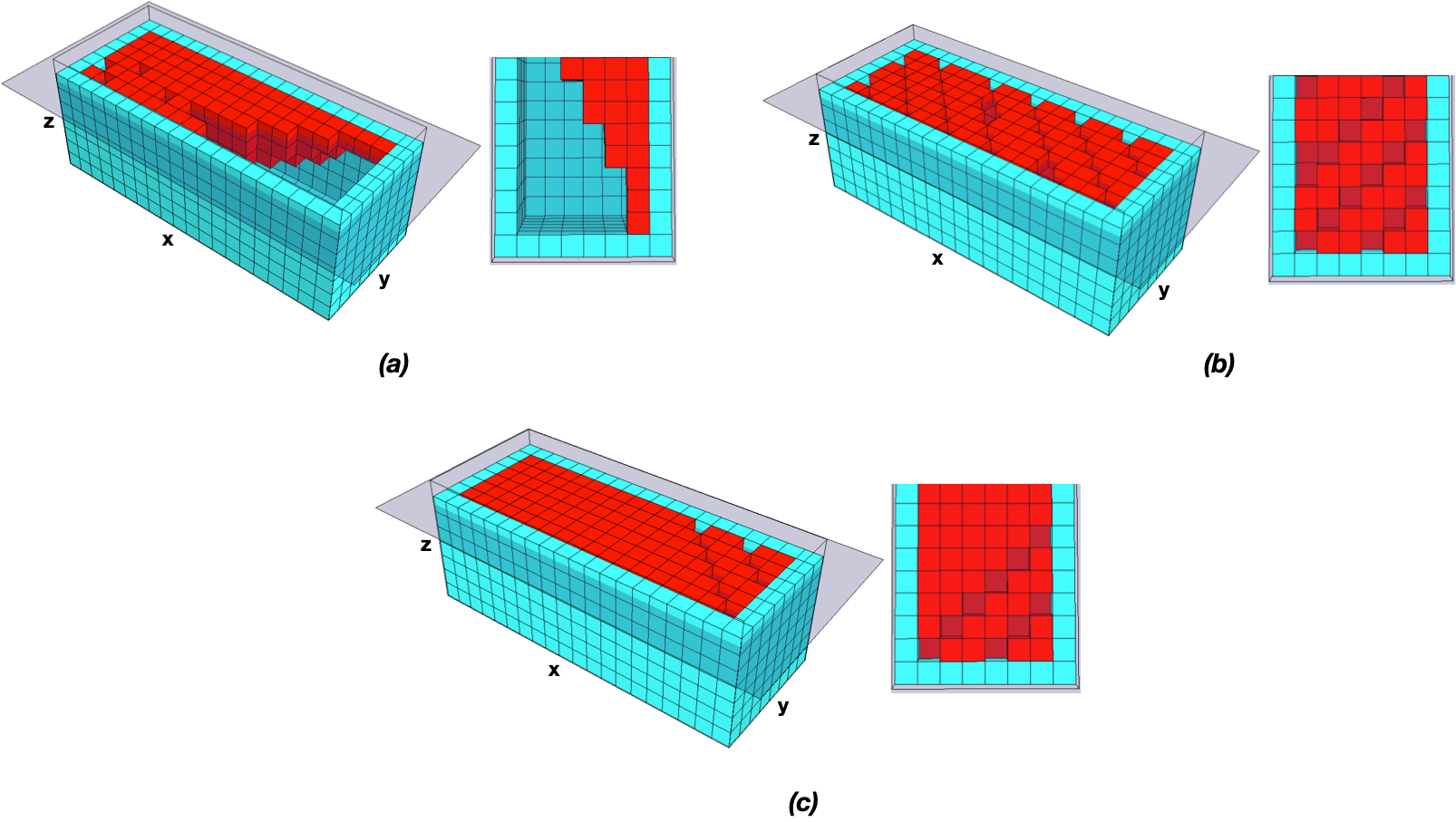}
  \caption{Morphology of the fittest actuator under: (a) AFPO; (b) NEAT; and (c) HyperNEAT. Blue voxels represent passive tissue; red voxels represent contractile tissue.}
  \label{fig:catheter_analysis_morphologies}
\end{figure}

For illustration reasons, Fig.~\ref{fig:catheter_analysis_morphologies} depicts the fittest actuator morphologies generated by all three approaches. The fittest actuator designed by AFPO (see Fig.~\ref{fig:catheter_analysis_morphologies}-a) exhibits a pyramidal pattern consisting of having a few contractile voxels at the bottom and gradually incrementing the number of voxels towards the top of the actuator. Arguably, the lack of contractile support at the bottom of the actuator affects its bending properties. Furthermore, the fittest actuator generated by NEAT (see Figure~\ref{fig:catheter_analysis_morphologies}-b) presents a striped diagonal pattern with no voxels throughout the contractile body composing the actuator. This pattern possibly allows the catheter actuator to bend suitably regardless of the controller phase offset. Finally, the fittest actuator created by HyperNEAT (see Figure~\ref{fig:catheter_analysis_morphologies}-c) also exhibits a striped diagonal pattern with no voxels, only in the free end of the actuator. The striped pattern allows the actuator to bend; however, the solid contractile mass on the other end of the actuator arguably attenuates the bending movement. 

{To sum up, the NE-based methodologies have produced the most robust actuator morphologies, that directly relates to a high chance of performing efficiently on real-world applications. The random phase offsets used in this study can be regarded to assess morphologies for noise resistance in the simulations that is a way to trivially predict the success of their physical counterpart when manufactured in a real-world application \cite{jakobi1997evolutionary}. The superior performance of AFPO on finding the fittest morphology can be preferred only in cases that the actuators can be controlled with high accuracy, due to the small variety of phase offsets tested throughout the evolutionary optimisation (for computational efficiency reasons). The proposed methodology can be instantiated with the parameters of the building blocks (or physical characteristics of available materials) to be utilised for the soft actuator manufacturing process, the physical constraints of the problem, the physical description of the environment that will accommodate the actuator and a target behavior (or fitness function). Then, through the evolutionary optimisation of the NE-based algorithms, an efficient morphology of the actuator will be produced, that will be in the form of what type of material should be placed at each point in the three-dimensional design space.}

\subsection{Error analysis}\label{sec:catheter_error}

{To understand the variability of the performance of actuators and, consequently, of the algorithms used in this research, two experiments are performed: (i) varying the properties of two simulated materials and (ii) altering the control strategy used in previous experiments. It is important to emphasize that the fittest actuator morphology found in the experiments described in Section~\ref{sec:catheter_analysis_robutsness} was utilised for the experiments described in the following sections. Namely, Morphology 9, found by NEAT (N-M9).}

\subsubsection{Varying material properties}\label{sec:catheter_error_parameters}

{This experiment changes the value of two mechanical properties of the contractile material used for the actuator morphologies during simulations. These two properties are:}

\begin{enumerate}
    \item {{\em Young's modulus}. It describes the stiffness of materials. This property is related to the elasticity of materials under compression or tension \cite{Jastrzebski1987}.}
    \item {{\em Poisson's ratio}. It is the negative ratio of transverse strain to axial strain. In other words, this property is the deformation of a material perpendicular to the load direction \cite{Jastrzebski1987}.}
\end{enumerate}

{During the previous experiments (see Section~\ref{sec:catheter}), Young's modulus and Poisson's ratio were set to {5e6} and 0.35, respectively. In this experiment, each value is varied in the following range: $[-10\%, -5\%, 5\%, 10\%]$.}

\begin{figure}[tb!]
  \centering
     \includegraphics[width=0.69\linewidth]{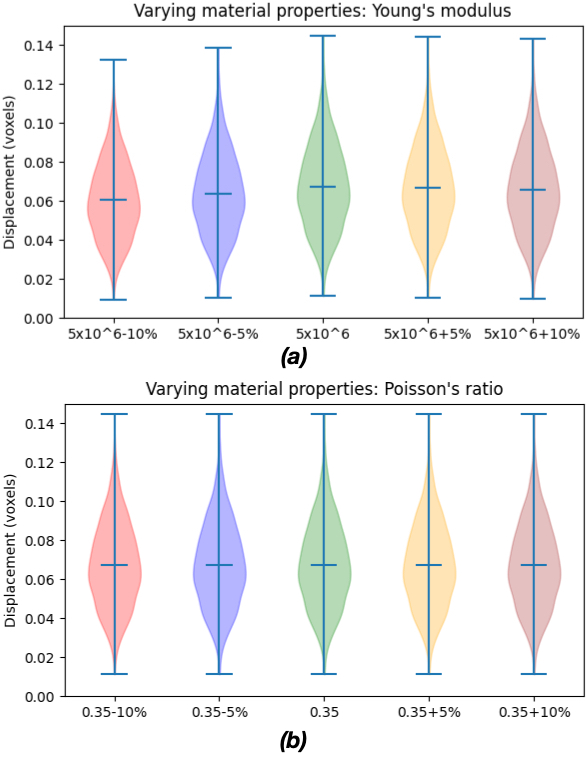}
  \caption{Performance of N-M9 varying the values of: (a) Young's modulus, and (b) Poisson's ratio.}
  \label{fig:catheter_error_parameters}
\end{figure}

{Figure~\ref{fig:catheter_error_parameters} depicts violin plots comparing the displacement exhibited (i.e., the performance) by N-M9 varying the values of the Young's modulus and the Poisson's ratio. Each violin plot shows the median, minimum, maximum and kernel density estimation of the frequency distribution under 1000 different offset controller scenarios.}

{When the value of Young's modulus is altered (see Figure~\ref{fig:catheter_error_parameters}-a), significant differences can be observed among the performances. All the data gathered are non-normally distributed (Shapiro-Wilk; $p<0.01$). It is possible to confirm significant differences between the performances whose Young's value is $<{5e6}$ (Dunn's test; $p<0.01$). Furthermore, significant differences are exhibited when these two performances are compared against those whose Young's value is $\geq {5e6}$ (Dunn's test; $p<0.01$). However, there are no significant differences among the performances whose Young's value is $\geq {5e6}$ (Dunn's test; $p>0.05$).}

{As expected, the results indicate that the material's stiffness significantly influences the actuator morphology's performance. In general, the less stiff the material is, the less controllable it is, and hence, the less performance is observed. On the other hand, if the material is stiffer, there is no significant improvement in performance.}

{Furthermore, when the value of the Poisson's ratio is varied, there are no significant differences among the performances. This is confirmed by first testing the collected data using the Shapiro-Wilk test and obtaining $p<0.01$, which indicates that the data is non-normally distributed. Then, applying Dunn's test, the result is $p>0.05$, suggesting that the deformation of the material does not significantly affect the performance of the actuator morphology.}

\subsubsection{Testing the control strategy}\label{sec:catheter_error_control}

{In this experiment, a random controller phase offset is used. This controller phase offset is copied 18 times. Each copy represents a scenario where a {\em slice} in the plane $yz$ of N-M9 is affected. The controller phase offset values associated with the voxels composing the slice are set to 0. It is essential to point out that although the dimension of N-M9 in the $x$ axis is 20, only 18 slices are available due to the passive enclosure embodied (see Section~\ref{sec:algorithms}). For visual reference, see  Figure~\ref{fig:catheter_analysis_morphologies}-b. Furthermore, the value of Young's modulus and Poisson's ratio are set to {5e6} and 0.35 respectively (see Section~\ref{sec:catheter_error_parameters}).}

\begin{figure}[tb!]
  \centering
     \includegraphics[width=0.67\linewidth]{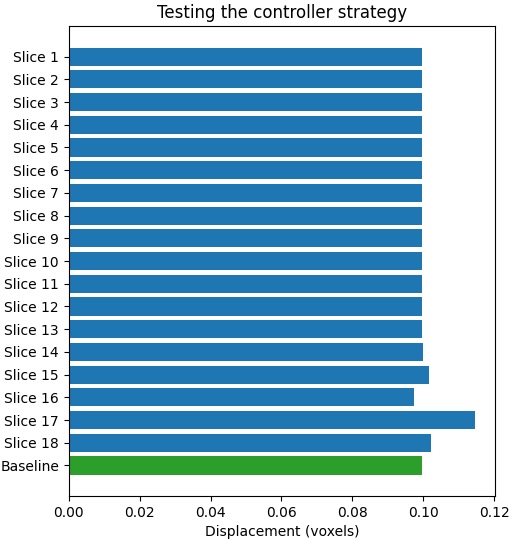}
  \caption{Performance of N-M9 under 18 different scenarios where the values of the controller phase offset associated with the $n$ slice are set to 0.}
  \label{fig:catheter_control_strategies}
\end{figure}

{Figure~\ref{fig:catheter_control_strategies} presents a bar plot showing the displacement reached by N-M9 when the values of the controller phase offset associated with $n$ slice are set to 0 (blue bars). In addition, a green bar shows the performance of N-M9 with no modification in the values of the controller phase offset; it is used as a baseline.}

{As expected, no significant variation in terms of performance is observed in the slices that are near the fixed end of N-M9. This behavior occurs from slices 1 to 14. Furthermore, those slices that are near the free end of N-M9 exhibit a significant variation, which is observed in slices 16 to 18.}

{In general, despite the specific noise applied to the controller input, the performance of the actuator morphology is consistent (i.e., there are no noticeable variations) in the vast majority of scenarios. On the other hand, the variations in performance observed are arguably due to external interactions that affect the material's mechanical properties, such as the absence of physical support (i.e., no fixed end).}


\section{Conclusions}\label{sec:conclusions}

This research studied the suitability of NEAT and HyperNEAT as generators of soft actuator morphologies. Their capabilities were compared against an implementation of AFPO, a popular multi-objective optimisation approach. The performance analysis was done utilising three metrics: (i) analyzing the general performance to find actuators with the maximum displacement possible in the $yz$ plane; (ii) testing the robustness of those actuators exhibiting the maximum displacement reached (i.e., the fittest actuators); and (iii) determining the fittest actuators with the minimum number of voxels possible. For this analysis, 20 different actuators were generated (i.e., 20 experimental runs were conducted) by each approach. 

Results suggest that neuroevolution-based approaches are generally more capable of producing robust catheter actuator morphologies than AFPO due to the genetic operators (i.e., crossover and mutation) induced a broader exploration in the search space. In addition, the actuator produced by NEAT could outperform the one generated by HyperNEAT. Arguably, the design of the substrate utilised during experimentation affected the performance of HyperNEAT.

Furthermore, the number of voxels plays a crucial role in bending tasks; the fittest actuator found by AFPO exhibited significantly fewer contractile voxels; however, its performance was not superior when considering the multitude of possible controller strategies. On the other hand, the fittest actuator generated by NEAT and HyperNEAT showed at least twice the number of voxels of the actuator generated by AFPO. Their performance, nevertheless, overcomes the one from the actuator generated by AFPO. Arguably, the striped no-voxel pattern exhibited by NEAT and HyperNEAT actuators helped to enhance the bending movement. {Although, the designs produced by the proposed NE-based methodologies were not validated \textit{in vitro}, this should not hinder their applicability or lessen their better performance demonstrated when compared with AFPO. The validation of AFPO applied on Voxelyze has been previously proved after fine-tuning the environment simulation parameters \cite{Kriegman2020}. The selection of materials to build the actuator are considered as the material parameters inputs to Voxelyze, while the manufacturing process of artificial material should not hinder the generalization of this methodology.} 

{The fact that the fitness function of the morphologies considered in the evolutionary optimisation methodology is provided by Voxelyze, a voxel-based physics simulator, may give rise to some concern around the translation of designs to real-world applications through concurrent manufacturing processes. Voxelyze utilises cube elements as building blocks and, thus, the morphologies produced may be predominately comprised by orthogonal boundaries; a characteristic that is not ideal for real-world applications. Namely, soft material (i.e., plastic melted through a 3D printer nozzle) or biohybrid material (i.e., living cells that are subject to tissue adherence/contraction) can not formulate perfectly right angles in the micro-scale. Nonetheless, the implementation of the evolutionary optimisation through indirect representations (i.e., CPPNs) can alleviate part of this limitation. These representations are scale free. Consequently, the anatomical resolution of the morphology may be increased without any further tests of the computational demanding evolutionary optimisation. The enhanced resolution may develop convex morphologies that would appear as orthogonal in lower resolutions \cite{Kriegman2020}. The alleviation of right angles can be further enhanced through the use of appropriate activation functions on NEAT algorithm. On the other hand, this is not true for all cases. Therefore, an implementation of filtering the morphologies for manufacturing compatibility can follow the \textbf{morphology generator} proposed here, in the automated design pipeline envisioned. These filters will remove morphologies that are not possible or trivial to build with the available manufacturing techniques. However, the implementation of these filters are considered as an aspect of future work.}

Considering the results and insights gathered from this research, future work may include the addition of more periodic activation functions, such as cosine and tangent. Following that tactic will help to induce more diverse morphological patterns produced by CPPNs. Another avenue of future work consists of a broader exploration of the number of hidden layers and neurons that may lead to better performance under HyperNEAT. In order to improve the design of the substrate, an approach called {\em ES-HyperNEAT}, focused on evolving the location of every neuron and the pattern of weights among them, can be implemented \cite{Risi2012}.

{Moreover, a more robust performance comparison against other NE-based approaches will be conducted. For instance, Evolutionary eXploration of Augmenting Memory Models (EXAMM), which is capable of evolving ANNs (including recurrent ANNs) through a wide variety of memory structures \cite{Ororbia2019,Thakur2023}. Taking into account that real-world optimisation problems can be time-constrained, the required computational time of further methodologies studied will be a key factor. Furthermore, the utilisation of surrogate-model assisted optimisation \cite{preen2019towards} will be considered, to minimize the need for computationally demanding physics simulators.} Finally, another crucial aspect of future work will be focused on evolving controllers for the fittest catheters actuators found in this research.


\section*{CRediT authorship contribution statement}

\textbf{Hugo Alcaraz-Herrera}: Data curation; Formal analysis; Investigation; Methodology; Software; Visualization; Writing - original draft; Writing - review \& editing
\textbf{Michail-Antisthenis Tsompanas}: Conceptualization; Funding acquisition; Methodology; Writing - review \& editing
\textbf{Igor Balaz}: Conceptualization; Funding acquisition; Project administration; Writing - review \& editing
\textbf{Andrew Adamatzky}: Conceptualization; Funding acquisition; Supervision; Writing - review \& editing

\section*{Declaration of Competing Interest}
The authors declare that they have no known competing financial interests or personal relationships that could have appeared to influence the work reported in this paper.

\section*{Acknowledgement}
This project has received funding from the European Union’s Horizon Europe research and innovation programme under grant agreement No. 101070328. UWE researchers were funded by the UK Research and Innovation grant No. 10044516.

\end{document}